\documentclass[journal]{IEEEtran}
\ifCLASSINFOpdf
\else
\fi
\AtBeginDocument{%
	\providecommand\BibTeX{{%
			\normalfont B\kern-0.5em{\scshape i\kern-0.25em b}\kern-0.8em\TeX}}}
\usepackage{times}
\usepackage{epsfig}
\usepackage{graphicx}
\usepackage{amsmath}
\usepackage{amssymb}
\usepackage[linesnumbered,boxed,ruled,commentsnumbered]{algorithm2e}
\usepackage{verbatim}
\usepackage{cite}
\begin{document}
%
\title{Diverse Video Captioning Through Latent Variable Expansion}
%
%
%

\author{Huanhou~Xiao,
        and~Jinglun~Shi
\thanks{The paper is under consideration at Pattern Recognition Letters.}}

%
%

\markboth{Journal of \LaTeX\ Class Files,~Vol.~14, No.~8, August~2015}%
{Shell \MakeLowercase{\textit{et al.}}: Bare Demo of IEEEtran.cls for IEEE Journals}
%



\maketitle

\begin{abstract}
Automatically describing video content with text description is challenging but important task, which has been attracting a lot of attention in computer vision community. Previous works mainly strive for the accuracy of the generated sentences, while ignoring the sentences diversity, which is inconsistent with human behavior. In this paper, we aim to caption each video with multiple descriptions and propose a novel framework. Concretely, for a given video, the intermediate latent variables of conventional encode-decode process are utilized as input to the conditional generative adversarial network (CGAN) with the purpose of generating diverse sentences. We adopt different Convolutional Neural Networks (CNNs) as our generator that produces descriptions conditioned on latent variables and discriminator that assesses the quality of generated sentences. Simultaneously, a novel DCE metric is designed to assess the diverse captions. We evaluate our method on the benchmark datasets, where it demonstrates its ability to generate diverse descriptions and achieves superior results against other state-of-the-art methods.
\end{abstract}


%
\IEEEpeerreviewmaketitle

\begin{figure*}[htp]
	\centering
	\includegraphics[width=\textwidth]{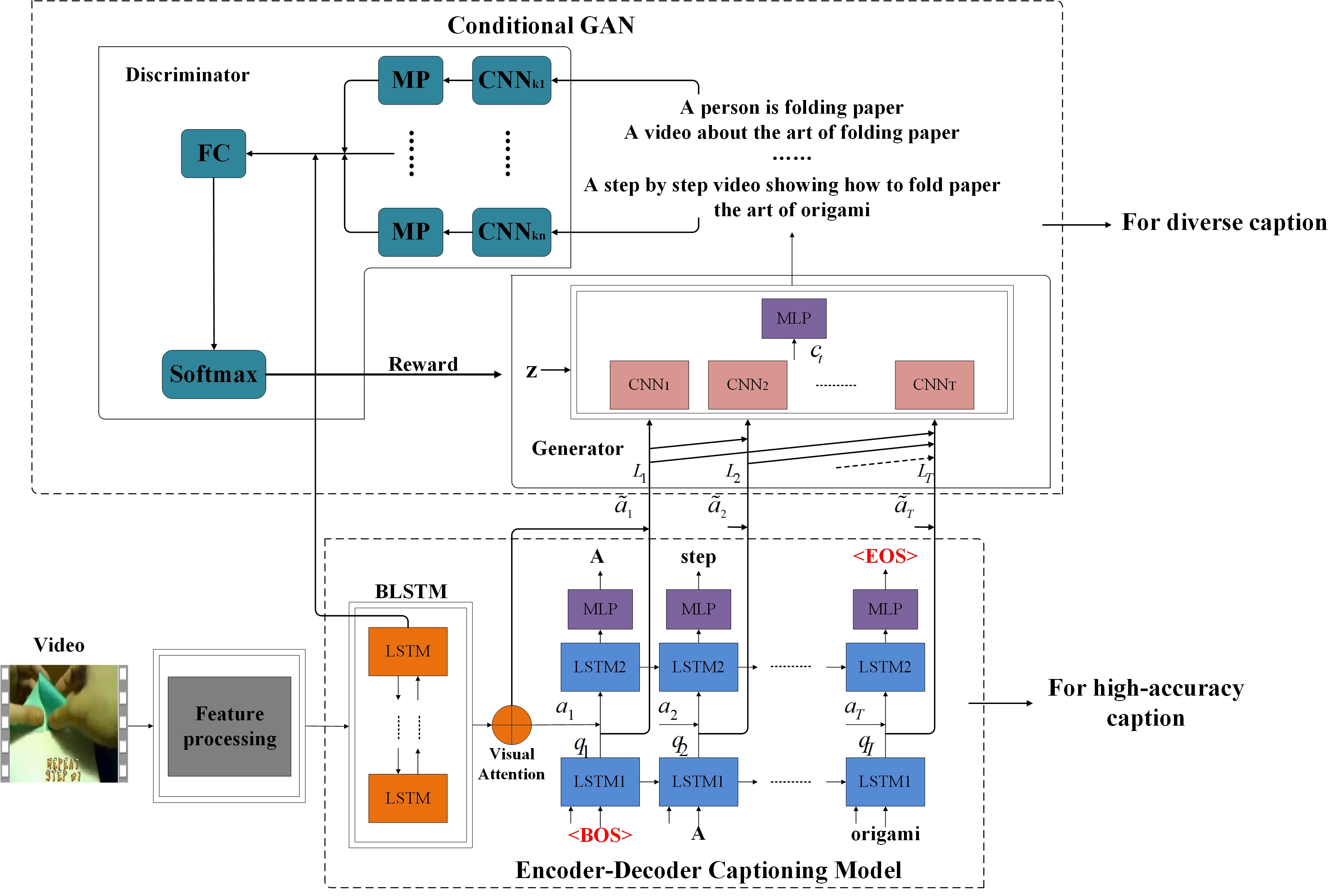}
	\caption{Illustration of the proposed method. Our model consists of an encoder-decoder architecture and a conditional GAN. The former aims to learn an effective coding method for video and text, and the latter is designed to generate diverse descriptions conditioned on the latent variables extracted from the encoder-decoder model. The design of hierarchical LSTM enables the encoder-decoder model to generate a single accurate sentence while also assisting the generator to generate diverse sentences. During testing, when we use LSTM2 to generate high-precision captions, the input of LSTM1 at each step is the previous generated word of LSTM2. And if we want to generate diverse descriptions, the input of LSTM1 is the previous generated word of generator. This allows our encoder-decoder model and generator to generate different sentences.
	}
	\label{fig:framework_singlefeature}
\end{figure*}
\section{Introduction}
%
%
%
%
Video captioning has recently received increased interest and become an important task in computer vision. Most of the research efforts have been trying to generate captions for videos, including the template based methods and the neural network based models. In the template based methods, subjects, verbs and objects are firstly detected and then filled into a pre-defined template to generate the corresponding sentence. However, the sentences yielded by these methods are very rigid and limited. They are still far from being perfect.

Deep learning has rapidly developed, and significant advancements have been made in image classification using convolutional neural network (CNN)~\cite{simonyan2014very} and machine translation utilizing recurrent neural network (RNN)~\cite{sutskever2014sequence}. Benefit from these achievements, neural network based approaches have become the mainstream to generate descriptions for videos. They are mostly built upon the encoder-decoder framework and optimized in an end-to-end manner. In these approaches, typically a CNN is utilized as an encoder to generate the visual representation and a RNN as a decoder to generate sequence of words.

In view of the effectiveness of the encoder-decoder framework, we follow this elegant recipe in our work. However, previous research primarily focuses on the fidelity of sentences, while another essential property, diversity, is not taken into account. More specifically, most models are trained to select words with maximum probability, which results in a monotonous set of generated sentences since these sentences bear high resemblance to training data.

Towards the goal of building a model that can selectively output a single high-accuracy description or the diverse descriptions, we propose a novel approach on top of the encoder-decoder framework and the conditional GAN, as depicted in Fig. 1. We refer to our proposed method as DCM (Diverse Captioning Model), which consists of two components: (1) An attention-based LSTM model is trained with cross entropy loss to generate textual descriptions for the given videos. Concretely, we employ a Bi-directional LSTM~\cite{graves2005framewise} to encode the video features. Following it, a temporal attention mechanism is utilized to make a soft-selection over them. Afterwards, we adopt the hierarchical LSTM to generate the descriptions. (2) A conditional GAN whose input is the latent variables of the attention-based model is trained to generate diverse sentences. Specifically, in CGAN, we get rid of LSTM and adopt the fully convolutional network as our generator to produce descriptions based on the latent variables. For the discriminator, both the sentences and video features are used as input to evaluate the quality of generated sentences. The former component is designed to effectively model video-sentence pairs, and the latter aims at simultaneously considering both the fidelity and diversity of the generated descriptions.

The contributions of this work can be summarized as follows: (1) A dual-stage framework that can generate both high-accuracy caption and diverse caption is proposed. Simultaneously, our diversity module is compatible with almost all models based on the encoder-decoder structure, as long as their decoders are simply extended to a hierarchical architecture. (2) We propose a novel performance evaluation metric named Diverse Captioning Evaluation (DCE), which considers not only the differences between sentences, but also the rationality of sentences (i.e., whether the video content is correctly described). (3) Extensive experiments and ablation studies on benchmark MSVD and MSR-VTT datasets demonstrate the superiority of our proposed model in comparison to the state-of-the-art methods. 

\section{Related Work}

\subsection{RNN for Video Captioning}

As a crucial challenge for visual content understanding, captioning tasks have attracted much attention for many years. Venugopalan \emph{et al.}~\cite{venugopalan2014translating} transferred knowledge from image caption models via adopting the CNN as the encoder and LSTM as the decoder. Pan \emph{et al.}~\cite{pan2016jointly} used the mean-pooling caption model with joint visual and sentence embedding. To better encode the temporal structures of video, Yao \emph{et al.}~\cite{yao2015describing} incorporated the local C3D features and a global temporal attention mechanism to select the most relevant temporal segments. More recently, a novel encoder-decoder-reconstruction network was proposed by~\cite{wang2018reconstruction} to utilize both the forward and backward flows for video captioning. Aafaq \emph{et al.}~\cite{Aafaq_2019_CVPR} embedded temporal dynamics in visual features by hierarchically applying Short Fourier Transform to CNN features.

Multi-sentence description for videos has been explored in several works~\cite{shin2016beyond,yu2016video,song2018deterministic,shen2017weakly}. Shin \emph{et al.}~\cite{shin2016beyond} temporally segmented the video with action localization and then generated multiple captions for those segments. A hierarchical model containing a sentence generator that produces short sentences and a paragraph generator that captures the inter-sentence dependencies was proposed by ~\cite{yu2016video}. DVC~\cite{shen2017weakly} exploited the spatial region information and further explored the correspondence between sentences and region-sequences. Different from these methods which focus on temporally segmentation or spatial region information, MS-RNN~\cite{song2018deterministic} modeled the uncertainty observed in the data using latent stochastic variables. It can thereby generate multiple sentences with consideration of different random factors. In this paper we also pay attention to the latent variables, but we try to generate diverse captions from the video-level.
\subsection{GAN for Natural Language Processing}

Generative adversarial network (GAN)~\cite{goodfellow2014generative} has become increasingly popular and achieved promising advancements in generating continuous data and discrete data~\cite{yu2017seqgan}. It introduces a competing process between a generative model and a discriminative model through a minimax game where the generative model is encouraged to produce highly imitated data and the discriminative model learns to distinguish them. To produce specific data, conditional GAN (CGAN) was first proposed by~\cite{mirza2014conditional} to generate MNIST digits conditioned on class labels. Since then it has been widely applied to other fields, such as image synthesis and text generation. For instance, Reed et al.~\cite{reed2016generative} used CGAN to generate images based on the text descriptions. Dai et al.~\cite{dai2017towards} built a conditional sequence generative adversarial net to improve the image description.

Motivated by previous researches, in this paper we incorporate CGAN with encoder-decoder model to describe video content. With this design, our proposed method is able to selectively output a single high-accuracy description or diverse descriptions.

\section{The Proposed Method}
\begin{figure}[t]
	\centering
	\includegraphics[width=\columnwidth]{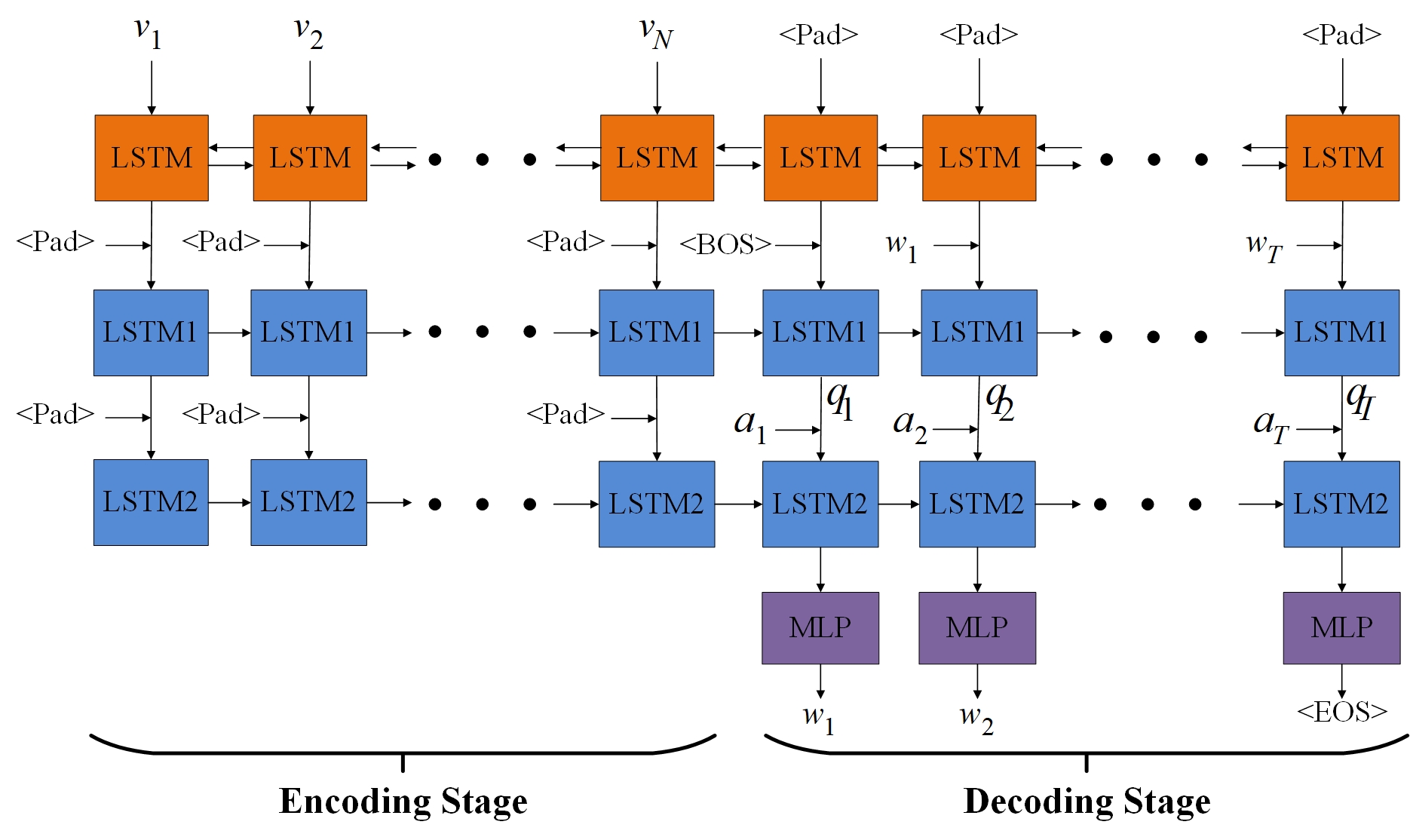}
	\caption{Framework of our unrolled encoder-decoder model. We use
		bi-directional LSTM to encode the visual features extracted from CNN, and use another LSTMs with temporal attention mechanism to generate the caption word sequence during the decoding stage.}
	\label{fig:Framework of our encoder-decoder model}
\end{figure}
\subsection{Single Captioning with Encoder-Decoder Framework}
We propose in this paper a stacked LSTM based model for single video captioning, as shown in Fig. 2. We refer to it Ours-ED. Given a video, \textbf{V}, with \emph{N} frames, the extracted visual features and the embedded textual features can be represented as $\textbf{v}=\{v_{1},v_{2},...,v_{N}\}$ and $\textbf{w}=\{w_{1},w_{2},...,w_{T}\}$, where $v_{i}$ $\in$ $\mathbb{R}^{D_{v}\times 1}$, $w_{i}$ $\in$ $\mathbb{R}^{D_{w}\times 1}$, and $T$ is the sentence length. Specifically, ${D}_{v}$ and ${D}_{w}$ are the respective dimensions of the frame-level features and vocabulary. We use a bi-directional LSTM (BLSTM) which can capture both forward and backward temporal relationships to encode the extracted visual features. The activation vectors are obtained as:
\begin{equation}
h_t=h^{(f)}_t+h^{(b)}_t
\end{equation}
where $h^{(f)}_t$ and $h^{(b)}_t$ are the forward and backward hidden activation vectors. Inspired by the recent success of attention mechanism in various tasks, a visual attention mechanism is incorporated following the BLSTM. To avoid imposing visual attention on non-visual words~\cite{song2017hierarchical}, we append a blank feature whose values are all zeros to the encoded video features. Thus, we can eliminate the impact of visual information if the predicting word is irrelevant to the high-level visual representations Accordingly, the output context vector at time step \emph{t} can be represented as:
\begin{equation}
a_t=\sum_{i=1}^{N+1}\alpha_{t,i}h_i
\end{equation}

In (2), $h_{N+1}$ is the blank feature, and $\alpha_{t,i}$ is the attention weight which can be computed as:
\begin{equation}
\alpha_{t,i}={\rm softmax}(e_{t,i})
\end{equation}
\begin{equation}
e_{t,i}=w^T{\rm tanh}(W_ah_i+V_ah^d_{t-1}+b_a)
\end{equation}
where $w$, $W_a$, $V_a$ and $b_a$ are the learned parameters, $h^d_{t-1}$ is the hidden state of the decoder LSTM (LSTM2) at the ($t$-1)-th time step.

Our decoder integrates two LSTMs. LSTM1 is used to efficiently encode the previous words and outputs $q_t$. Then LSTM2 generates the next word based on the concatenation of the visual information $a_t$ and the textual information $q_t$. With this hierarchical design, our model can not only generate a concise and accurate description, but also assist the diversity module to generate varied descriptions, which will be introduced in the following section. According to the above analysis, at time step $t$, our model utilizes \textbf{V} and the previous words $w_{<t}$ to predict a word $w_{t}$ with the maximal probability $P(w_t|w_{<t},\textbf{V})$, until we reach the end of the sentence. Thus, the loss function of our encoder-decoder model can be defined as:
\begin{equation}
L_{xe}=-\sum_{t=1}^T{\rm log}P(w_t|w_{<t},\textbf{V};\theta)
\end{equation}
where $\theta$ is the model parameter set.

\subsection{Diverse Captioning with Conditional GAN}

As introduced above, we use the conditional GAN to implement the diverse captioning for video. In our framework, multiple CNNs are adopted as the generator G to generate descriptions, and the discriminator D trying to assess the sentences quality. They are optimized by a minimax two-player game until they reach an equilibrium.

Concretely, our generator strives to produce high-quality sentences conditioned on the random vector \textbf{z} sampled from the normal distribution $\mathcal{N}$(0,1) and the latent variables \textbf{L} obtained from the encoder-decoder model, where $\textbf{z}=\{z_{1},z_{2},...,z_{T}\}$ and $\textbf{L}=\{L_{1},L_{2},...,L_{T}\}$. Among them, $L_{t}$ consists of $\widetilde{a}_t$ and $q_t$, and $\widetilde{a}_t$ is calculated in the same way as $a_t$, except that the output of the ${\rm CNN_t}$ is used as input. At each step $t$, suppose $s_t$ is the concatenation of $z_t$ and $L_t$, and $S_{1:t}\in \mathbb{R}^{t\times \bar{k}}=(s_1,s_2,...,s_t)$ is the concatenation of $s_{i,\,i\in [1,t]}$, where $\bar{k}$ is the dimension of $s_{t}$. Then, a ${\rm CNN_t}$ with kernel $K_c^g$ $\in$ $\mathbb{R}^{t \times \bar{k}}$ takes input $S_{1:t}$ and produces a context vector $C_t$ that contains the information of all the previously generated words. Afterwards, a MLP layer is utilized to encode $C_t$ and yields a conditional distribution $P(w_t|z_t,L_{1:t})$ over the vocabulary. We choose CNN since it is being penalized for producing less-peaky word probability distributions, giving it the potential to explore the sentence diversity~\cite{aneja2018convolutional}. Overall, the random vector allows the generator to generate diverse descriptions and the latent variable guarantees the fidelity of the generated sentences. Specifically, when the discriminator is fixed, the generator can be optimized by minimizing the following formulation:
\begin{equation}
\min\limits_{\phi}\mathbb{E}_{\textbf{z}\sim\mathcal{N}}[{\rm log}(1-D(G_{\phi}(\textbf{z},\textbf{L})))]
\end{equation}
here, $\phi$ represents the parameter set of generator G. Considering both the accuracy and diversity of the generated sentences, we balance the generator with an extra cross entropy loss, which also prevents our conditional GAN from deviating from its correct orbit during training. Therefore, we minimize the objective function $L_G$ as follows when updating G:
\begin{equation}
\begin{aligned}
L_G=&-\lambda\mathbb{E}[{\rm log}P(\textbf{w}|\textbf{z},\textbf{L})]+(1-\lambda)\mathbb{E}_{\textbf{z}\sim\mathcal{N}}[{\rm log}(1-D(\\&G_\phi(\textbf{z},\textbf{L})))]
\end{aligned}
\end{equation}
where $\lambda$ is the tradeoff parameter.  

In the discriminator D, our goal is to judge whether the given video-text pair as matched or not and how well a sentence describes a given video. Inspired by the superior performance of convolutional neural network in text classification, we also choose CNN as our discriminator. Suppose the embedded matrix of the generated sentence is represented as $X_{1:T}$ $\in$ $\mathbb{R}^{T\times k}=(x^1,x^2,...,x^T)$ by concatenating the word embeddings as columns, where $k$ is the dimension of the word embedding. Then, a convolutional operation with a kernel $K_c^d$ $\in$ $\mathbb{R}^{l \times k}$ is used to encode the sentence and produce a feature map:
\begin{equation}
f_i=g(K_c^d*X_{i:i+l-1}+b)
\end{equation}
here, * is the convolution operator, $b$ is the bias term and $g(\cdot)$ is the non-linear function (we use ReLU in our experiments). A max-over-time pooling operation is then applied over the generated feature maps, $\widetilde{f}$=$max\{f_1,f_2,...,f_{T-l+1}\}$. Moreover, we use different kernels to extract different features of sentence and combine them as our final sentence representation.

Once the above operations are completed, we concatenate the sentence representation with its corresponding video feature $h_N$ extracted from the hidden layer of BLSTM. Then, a fully connected layer is utilized to map this concatenated feature $H$ to a low-dimensional space and a softmax layer is incorporated to output the probability that indicates the quality of sentence. Formally, this process can be described as:
\begin{equation}
D(X)={\rm softmax}(W_pH+b_p)
\end{equation}
where $W_p$ and $b_p$ are parameters to be learned, $D(X)$ is projected onto the range of $[0,1]$. An output close to 1 indicates a bigger probability that $X$ is drawn from the real data distribution or not. The optimization target of D is to maximize the probability of correctly distinguishing the ground truth from the generated sentences. For G fixed, D can be optimized as:
\begin{equation}
\max\limits_{\eta}\mathbb{E}_{x\sim{P_{data}}}[{\rm log}(D_{\eta}(x)]+\mathbb{E}_{\textbf{z}\sim\mathcal{N}}[{\rm log}(1-D_{\eta}(G(\textbf{z},\textbf{L})))]
\end{equation}
where $\eta$ is the parameter set of discriminator D, $x$ represents the video-sentence pair in the training set $P_{data}$.

Unlike a typical GAN setting, in which the generator receives rewards at each intermediate step, our captioning model only receives the reward at the end (the reward signal is meaningful only for the completed sentence), which may lead to several difficulties in training such as vanishing gradients and error propagation. To mitigate the lack of intermediate rewards, the Monte Carlo rollouts~\cite{liu2017improved} is employed to provide early feedback.  Moreover, we model the captioning process via policy gradient~\cite{sutton2000policy} to overcome the problem that gradients can not be back-propagated directly since the text generation procedure is non-differentiable. Specifically, following~\cite{sutton2000policy}, when there is no intermediate reward, the objective of the generator (when optimizing the adversarial loss) is to generate a sequence $W_{1:T}=(w_1^,,w_2^,,...,w_T^,)$ from the start state $s_0$ to maximize its expected end reward:
\begin{equation}
\mathbb{E}[R_T|s_0,\phi]=\sum_{w_1^,\in\gamma}G_{\phi}(w_1^,|s_0) \cdot Q_{G_{\phi}}(s_0,w_1^,)
\end{equation}
\IncMargin{1em} 
\begin{algorithm}[t]
	\SetAlgoNoLine 
	\SetKwInOut{Input}{\textbf{Require}}
	\Input{
		encoder-decoder model $\pi_{\theta}$; generator $G_{\phi}$; discriminator $D_{\eta}$; training data set $P_{data}$}
	\BlankLine
	Initialize $\pi_{\theta}$, $G_{\phi}$ and $D_{\eta}$ with random weights $\theta$, $\phi$ and $\eta$\; 
	Pre-train $\pi_{\theta}$ on $P_{data}$ by Eq. (5)\;
	Generate negative samples in three ways mentioned in section III-B\;
	Pre-train $D_{\eta}$ via minimizing the cross entropy\;
	\Repeat
	{\textit{DCM converges}}
	{
		\For {g-step}{
			Generate a sequence $W_{1:T}=(w_1^,,w_2^,,...,w_T^,)$ using $G_\phi$\;
			\For {t in 1:$T$}{
				Compute $Q(s=W_{1:t-1},a=w_t^,)$ by Eq. (13)\;
			}
			Update generator parameters by Eq. (7)\;
		}
		\For {d-steps}{
			Use current $G_\phi$ to generate negative examples and combine with given positive examples of $P_{data}$\;
			Train discriminator $D_\eta$ for $p$ epochs by Eq. (10)\;
		}   
	}
	\caption{Training process of DCM}
\end{algorithm}
\DecMargin{1em}
where $s_0$ is the start state, $R_T$ is the reward for a complete sequence, $\gamma$ is the vocabulary, $G_{\phi}$ is the generator policy which influences the action that generates the next word, and $Q_{G_{\phi}}(s,a)$ indicates the action-value function of a sequence (i.e. the expected accumulative reward starting from state $s$, taking action $a$, and then following policy $G_{\phi}$). In our experiments we use the estimated probability of being real by the discriminator $D_{\eta}(W_{1:T},\hat{v})$ as the reward. Thus, we have:
\begin{equation}
Q_{G_{\phi}}(s=W_{1:T-1},a=w_T^,)=D_{\eta}(W_{1:T},\hat{v})
\end{equation}

Here, $\hat{v}$ indicates the corresponding visual feature. To evaluate the action-value for an intermediate state, the Monte Carlo rollouts is utilized to sample the unknown last $T-t$ tokens. To reduce the variance, we run the rollout policy starting from current state till the end of the sequence for \emph{K} times. Thus, we have:

$Q_{G_{\phi}}(s=W_{1:t-1},a=w_t^,)=$
\begin{equation}
\left\{
\begin{array}{lr}
\frac{1}{K}\sum\nolimits_{n=1}^KD_{\eta}(W_{1:T}^n,\hat{v}),W_{1:T}^n \in {\rm MC}(W_{1:t};K)&\textit{for t$<$\emph{T}}\\
D_{\eta}(W_{1:t},\hat{v})&\textit{for t$=$\emph{T}} 
\end{array}
\right.
\end{equation}
where $W_{1:t}^n=(w_1^,,...,w_t^,)$ and $W_{t+1:T}^n$ is sampled based on the rollout policy and the current state. In summary, Algorithm 1 shows full details of the proposed DCM. Aiming at reducing the instability in training process, we pre-train our encoder-decoder model and discriminator aforementioned to have a warm start. When pre-training our discriminator, the positive examples are from the given dataset, whereas the negative examples consist of two parts. One part is generated from our generator, and the other part is manually configured. Concretely, mismatched video-sentence pairs are utilized as one of the negative examples (model the inter-sentence relationship). Meanwhile, with the purpose of evaluating sentences more accurately, we exchange the word positions of the sentences in the positive examples and regard them as negative examples (model the intra-sentence relationship). The objective function of D for pre-training can be formalized into a cross entropy loss as follow:
\begin{equation}
L_D=-\frac{1}{m}\sum_{i=1}^{m}[Y_i{\rm log}(D(\textbf{w}_i))+(1-Y_i)({\rm log}(1-D(\textbf{w}_i)))]
\end{equation}
where $Y_i$ and $D(\textbf{w}_i)$ denote the real label and the predicted value of discriminator respectively, and $m$ is the number of examples in a batch. It is worth noting that during testing, when we use LSTM2 to generate high-precision captions, the input of LSTM1 at each step is the previous generated word of LSTM2. And if we want to generate diverse descriptions, the input of LSTM1 is the previous generated word of generator.

\begin{table}[t]
	\begin{center}
		\caption{Comparison with state-of-the-art methods on MSVD dataset. (-) indicates such metric is unreported.} \label{tab:msvd}
		\begin{tabular}{ c c c c c}
			\hline
			Method & METEOR & BLEU-4 & CIDEr & ROUGE-L
			\\ 
			\hline
			S2VT~\cite{venugopalan2015sequence} & 29.8 & - & - & - \\
			LSTM-E~\cite{pan2016jointly} & 31.0 & 45.3 & - & -\\
			h-RNN~\cite{yu2016video} & 32.6 & 49.9 & - & -\\
			HRNE~\cite{pan2016hierarchical} & 33.9 & 46.7 & - & -\\
			LSTM-TSA~\cite{Pan2017Video}& 33.5 & 52.8 & 74.0 & -\\
			aLSTMs~\cite{gao2017video} & 33.3 & 50.8 & 74.8 & -\\
			MS-RNN~\cite{song2018deterministic} & 33.8 & 53.3 & 74.8 & 70.2\\
			hLSTMat~\cite{song2017hierarchical} & 33.6 & 53.0 & 73.8 & -\\
			RecNet~\cite{wang2018reconstruction} & 34.1 & 52.3 & 80.3 & 69.8\\
			HMM~\cite{wang2018hierarchical} & 33.8 & 52.9 & 74.5 & -\\
			GRU-EVE~\cite{Aafaq_2019_CVPR}  &35.0 & 47.9 & 78.1 & \textbf{71.5}\\
			\hline
			Ours-ED & \textbf{35.6} & \textbf{53.3} & \textbf{83.1} & 71.2\\
			\hline
		\end{tabular}
	\end{center}
\end{table}
\begin{table}[t]
	\begin{center}
		\caption{Comparison with state-of-the-art methods on MSR-VTT dataset. (-) indicates such metric is unreported.} \label{tab:msrvtt}
		\begin{tabular}{c c c c c}
			\hline
			Method & METEOR & BLEU-4 & CIDEr & ROUGE-L
			\\ 
			\hline
			LSTM-GAN~\cite{yang2018video}  & 26.1 & 36.0 & - & -\\
			aLSTMs~\cite{gao2017video}  & 26.1 & 38.0 & 43.2 & -\\
			MS-RNN~\cite{song2018deterministic} & 26.1 & 38.8 & 40.9 & 59.3\\
			hLSTMat~\cite{song2017hierarchical} & 26.3 & 38.3 & - & -\\
			RecNet~\cite{wang2018reconstruction} & 26.6 & 39.1 & 42.7 & -\\
			DVC~\cite{shen2017weakly}  & 28.3 & 41.4 & \textbf{48.9} & 61.1 \\
			MARN~\cite{pei2019memory}  & 28.1 & 40.4 & 47.1 & 60.7\\
			OA-BTG~\cite{zhang2019object}  & 28.2 & 41.4 & 46.9 & -\\
			CVC-PSG~\cite{wang2019controllable}  & 28.2 & 42.0 & 48.7 & \textbf{61.6}\\
			\hline
			Ours-ED & \textbf{28.7} & \textbf{43.4} & 47.2 & \textbf{61.6}\\
			\hline
		\end{tabular}
	\end{center}
\end{table}
\section{Experiments}
\subsection{Datasets}
Two benchmark public datasets including MSVD~\cite{chen2011collecting} and MSR-VTT~\cite{xu2016msr} are employed to evaluate the proposed diverse captioning model. The MSVD dataset consists of 1,970 video clips collected from YouTube. It covers a lot of topics and is well-suited for training and evaluating a video captioning model. We adopt the same data splits as provided in~\cite{venugopalan2015sequence} with 1,200 videos for training, 100 videos for validation and 670 videos for testing. As for MSR-VTT, there are 10K video clips, each of which contains 20 reference sentences annotated by human. We follow the public split method: 6,513 videos for training, 497 videos for validation, and 2,990 videos for testing.
5
\subsection{Experimental Settings}

We uniformly sample 60 frames from each clip and use Inception-v3~\cite{szegedy2016rethinking} to extract frame-level features. To capture the video temporal information, the C3D network~\cite{karpathy2014large} is utilized to extract the dynamic features of video. The dynamic features are then encoded by a LSTM whose final output is concatenated with all the frame-level features. For MSR-VTT, we integrate the audio features extracted from the pre-trained VGGish model~\cite{hershey2017cnn} with other video features. We convert all the sentences to lower cases, remove punctuation characters and tokenize the sentences. We retain all the words in the dataset and thus obtain a vocabulary of 13,375 words for MSVD, 29,040 words for MSR-VTT. To evaluate the performance of our model, we utilize METEOR~\cite{denkowski2014meteor}, BLEU~\cite{papineni2002bleu}, CIDEr~\cite{vedantam2015cider}, and ROUGE-L~\cite{lin2004rouge} as our evaluation metrics, which are commonly used for performance evaluation of video captioning methods.

In our experiments, with an initial learning rate $10^{-5}$ to avoid the gradient explosion, the LSTM unit size and word embedding size are set as 512, empirically. We train our model with mini-batch 64 using ADAM optimizer~\cite{kingma2014adam}, and the length of sentence \emph{T} is set as 25. For sentence with fewer than 25 words, we pad the remaining inputs with zeros. To regularize the training and avoid overfitting, we apply dropout with rate of 0.5 on the outputs of LSTMs. In the training process, we first pre-train the encoder-decoder model and discriminator, and then optimize the CGAN module. When training CGAN, we freeze the parameters of the encoder-decoder model. During testing, beam search with beam width of 5 is used to generate descriptions.
\begin{table*}[t]
	\begin{center}
		\caption{Diversity evaluation using multiple metrics.} \label{tab: Diversity evaluation using multiple metrics}
		\resizebox{\textwidth}{!}{
			\begin{tabular}{c|c c c c c | c c c c c}
				\hline
				Dataset &\multicolumn{5}{c}{MSVD} & \multicolumn{5}{c}{MSR-VTT}\\
				\hline
				Model & DCE & DCE-Variant & Word Types & Mixed mBLEU & Self-CIDEr &  DCE & DCE-Variant & Word Types & Mixed mBLEU & Self-CIDEr\\
				\hline
				\hline
				MS-RNN~\cite{song2018deterministic} & 0.09 & 0.12 & 239 & 0.91 & 0.26 & 0.13 & 0.21 & 620 & 0.81 & 0.48\\
				DCM & \textbf{0.25} & \textbf{0.37} & \textbf{2210} & \textbf{0.64} & \textbf{0.79} & \textbf{0.21} & \textbf{0.39} & \textbf{6492} & \textbf{0.68} & \textbf{0.81}\\
				\hline
		\end{tabular}}
	\end{center}
\end{table*}

\subsection{Comparison with State-of-the-Art Models}
\subsubsection{Quantitative Analysis}
\textbf{Single Caption Evaluation.} Table I demonstrates the result comparison among our proposed method and some state-of-the-art models. The comparing algorithms include the encoder-decoder based architectures (S2VT~\cite{venugopalan2015sequence}, LSTM-E~\cite{pan2016jointly}, LSTM-TSA~\cite{Pan2017Video}, MS-RNN~\cite{song2018deterministic}, GRU-EVE~\cite{Aafaq_2019_CVPR}), and the attention based methods (h-RNN~\cite{yu2016video}, HRNE~\cite{pan2016hierarchical}, aLSTMs~\cite{gao2017video}, hLSTMat~\cite{song2017hierarchical},  HMM~\cite{wang2018hierarchical}, RecNet~\cite{wang2018reconstruction}). For MSVD dataset, the results of Ours-ED indicate that our method outperforms previous video captioning models on most metrics. In particular, compared with the best counterpart (i.e., GRU-EVE), Ours-ED achieves better performance, with 0.6$\%$, 5.4$\%$, and 5.0$\%$ increases on METEOR, BLEU-4, and CIDEr respectively, verifying the superiority of our proposed approach.

For MSR-VTT dataset, we also compare our models with some recent state-of-the-art methods~\cite{pei2019memory,zhang2019object,wang2019controllable}. MARN~\cite{pei2019memory} designs a memory structure to explore the full-spectrum correspondence between a word and its various similar visual contexts across videos. OA-BTG~\cite{zhang2019object} captures detailed temporal dynamics for salient objects and learns discriminative spatio-temporal representations based on object-aware aggregation with bidirectional temporal graph. CVC-PSG~\cite{wang2019controllable} guides the video caption generation with Part-of-Speech information and constructs a novel gated fusion network to fuse different types of video representations. The results are reported in Table II. Note that DVC adopts data augmentation during training to obtain better accuracy. Nevertheless, Ours-ED also achieves the best performance on METEOR, BLEU-4, and ROUGE-L. It demonstrates the effectiveness of our proposed method.

\begin{table}[t]
	\begin{center}
		\caption{Correlation to human evaluation.} \label{tab:Correlation to human evaluation}
		\begin{tabular}{c c c c}
			\hline
			Model & DCE-Variant & DCE & Human Evaluation\\
			\hline
			Ours-ED & 0.30 & 0.19 & 3.14 \\
			DCM & 0.39 & 0.26 & 3.92 \\
			GT & \textbf{0.53} & \textbf{1.05} & \textbf{5.96}\\
			\hline
		\end{tabular}
	\end{center}
\end{table}
\begin{table}[t]
	\begin{center}
		\caption{Diversity comparison on MSR-VTT.} \label{tab:Diversity comparison with DVC }
		\begin{tabular}{c c}
			\hline
			Model & DCE\\
			\hline
			AG-CVAE~\cite{wang2017diverse} & 0.18 \\
			DVC~\cite{shen2017weakly} & 0.19 \\
			DCM & \textbf{0.21} \\
			\hline
		\end{tabular}
	\end{center}
\end{table}
\begin{figure*}[t]
	\centering
	\includegraphics[width=\textwidth,height=5.5cm]{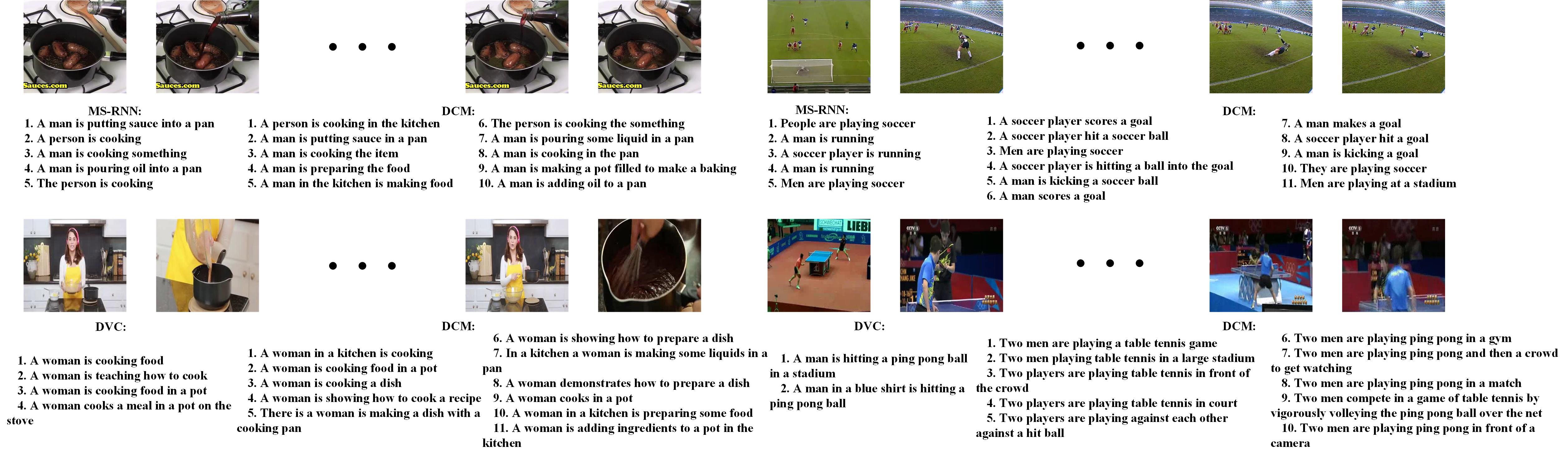}
	\caption{Examples of video captioning results. The results of MS-RNN and DVC are reported in~\cite{song2018deterministic} and~\cite{shen2017weakly}}.
	\label{fig:picture_4videos}
\end{figure*}
\textbf{Diversity Evaluation.} To evaluate the caption diversity quantitatively,~\cite{shen2017weakly} pay attention to its opposite - the similarity of captions. They uses latent semantic analysis (LSA)~\cite{deerwester1990indexing} to represent sentence by first generating sentence bag-of-words representation and then mapping it to LSA space. However, this method suffers important limitations. On one hand, it ignores the rationality of sentences. Two very different sentences can get high diversity score using this method, but their descriptions may be wrong. The generated sentences should be diverse on the premise of correctly describing the video content. On the other hand, this LSA method cannot capture polysemy and ignores the order of words in sentences, which limits its representation of sentence structure. Therefore, we propose in this subsection a diverse captioning evaluation (DCE) metric. It is mainly designed from two perspectives: the difference between sentences and the sentences reasonableness. About the difference of sentences, different from~\cite{shen2017weakly}, we use the Jaccard similarity coefficient~\cite{real1996probabilistic} which is effective in dealing with discrete data to model the word-level similarity, and use BERT~\cite{devlin2018bert} which shows superior performance in various NLP tasks and addresses the limitations of bag-of-words model to generate sentence-level representation. Meanwhile, we use METEOR as an assessment of sentences reasonableness since it has shown better correlation with human judgment compared with other metrics. Formally, the DCE can be calculated as:
\begin{equation}
\begin{aligned}
{\rm DCE}= &\frac{1}{m_v}\sum_{k=1}^{m_v}\frac{1}{n}\sum_{s^i,s^j\in \textbf{S},i\neq j}[M(s^i)+M(s^j)]\cdot [\delta(1-\\&J(s^i,s^j))+(1-\delta)(1-<BT(s^i),BT(s^j)>)]
\end{aligned}
\end{equation}
where $\textbf{S}$ is the sentence set with cardinality $n$ (i.e. if each video is accompanied with 5 generated sentences, $n$ will be 10 because each sentence is used to calculate the similarities with others), $m_v$ is the number of videos in the dataset, $\delta$ is the adjustment coefficient (we set $\delta=0.5$ in our experiments), $M(s)$ is the METEOR score of candidate sentence $s$, $BT(s)$ is the sentence vector encoded by BERT, $J$ and $<>$ denote the Jaccard similarity and cosine similarity.

We evaluate the diversity degree of our generated captions on the test set of two benchmark datasets. In addition to DCE, we also use the following methods: 
(1) \textbf{DCE-Variant}. The proposed DCE metric combines both accuracy and diversity. To interpret the role of each part more clearly, we evaluate the captions with the variant of DCE that removes the METEOR score. Thus we can see how much of the score comes from the sentence diversity. 
(2) \textbf{Word Types}. We measure the number of word types in the generated sentences. This reflects from the side whether the generated sentences are diverse. In general, higher word types indicates higher diversity.
(3) \textbf{Mixed mBLEU}. Given a set of captions, BLEU score is computed between each caption against the rest. Mean of these BLEU scores is the mBLEU score. Following~\cite{wang2019describing}, we consider the mixed mBLEU score which represents the mean of mBLEU-n, where n is the n-gram. A lower mixed mBLEU value indicates higher diversity. 
(4) \textbf{Self-CIDEr}. To measure the diversity of image captions, Self-CIDEr~\cite{wang2019describing} is designed to consider phrases and sentence structures. It shows better correlation with human judgement than the LSA-based metric. Higher is more diverse.

The results are shown in Table III, from which we observe that our DCM outperforms MS-RNN on all the evaluation metrics, showing the superiority of our proposed method. At the same time, it also proves the effectiveness of DCE which performs consistent with other metrics (e.g. Self-CIDEr). We observe that the values of our DCE are lower than DCE-Variant. This is because DCE penalizes the inaccurate descriptions. Different from other existing metrics, DCE is the first one that considers both diversity and accuracy of captions, which is vital for video captioning task. Because the diversity of captions is meaningful only when the video content is accurately described. At the same time, DCE evaluates the diversity from the word level and sentence level, and uses the advanced model BERT in NLP field to encode sentences, so the results are more accurate and reasonable.

To further verify the rationality of DCE, we also conduct human evaluation by inviting volunteers to assess the captions obtained by Ours-ED, DCM and Ground Truth on the test set of MSR-VTT. Note that Ours-ED also can generate multiple descriptions using beam search. However, with beam search-n, Ours-ED can only generate up to n different sentences for the same model, while DCM can generate more, even several times Ours-ED.  We randomly select 100 videos and sample 5 sentences from each model per video since Ours-ED has at most 5 different sentences. Each video is evaluated by 5 volunteers, and the average score is obtained. The given score (from 1 to 7) reflects the diversity of the set of captions. The higher the score, the better the diversity. From Table IV we can see that the DCE evaluation is consistent with the human evaluation, showing the advantage of DCE and proving that DCM is better than Ours-ED in diversity. Besides, compared with DCE-Variant, we notice that the gap between DCM and Ground Truth is larger under DCE evaluation. This indicates DCE takes both accuracy and diversity into account. Our DCE can be further improved if there is a better accuracy assessment method.

We also compare our method with DVC~\cite{shen2017weakly} and AG-CVAE~\cite{wang2017diverse} on the test set of MSR-VTT in Table V. AG-CVAE proposes a simple novel conditioning mechanism with an additive
Gaussian prior to structure the latent space in a way that can directly reflect object co-occurrence. We observe that our DCM achieves the best performance in terms of DCE (the LSA score reported in~\cite{shen2017weakly} is 0.501, and we also surpass it (Ours is 0.53)). In addition, observing the fact that some of the expressions generated by our model do not even appear in the reference sentences, we believe that it reflects the potential of our model to enrich the annotation sentences in the database, which is of great significance.

\begin{figure*}[t]
	\centering
	\includegraphics[width=\textwidth,height=16.5cm]{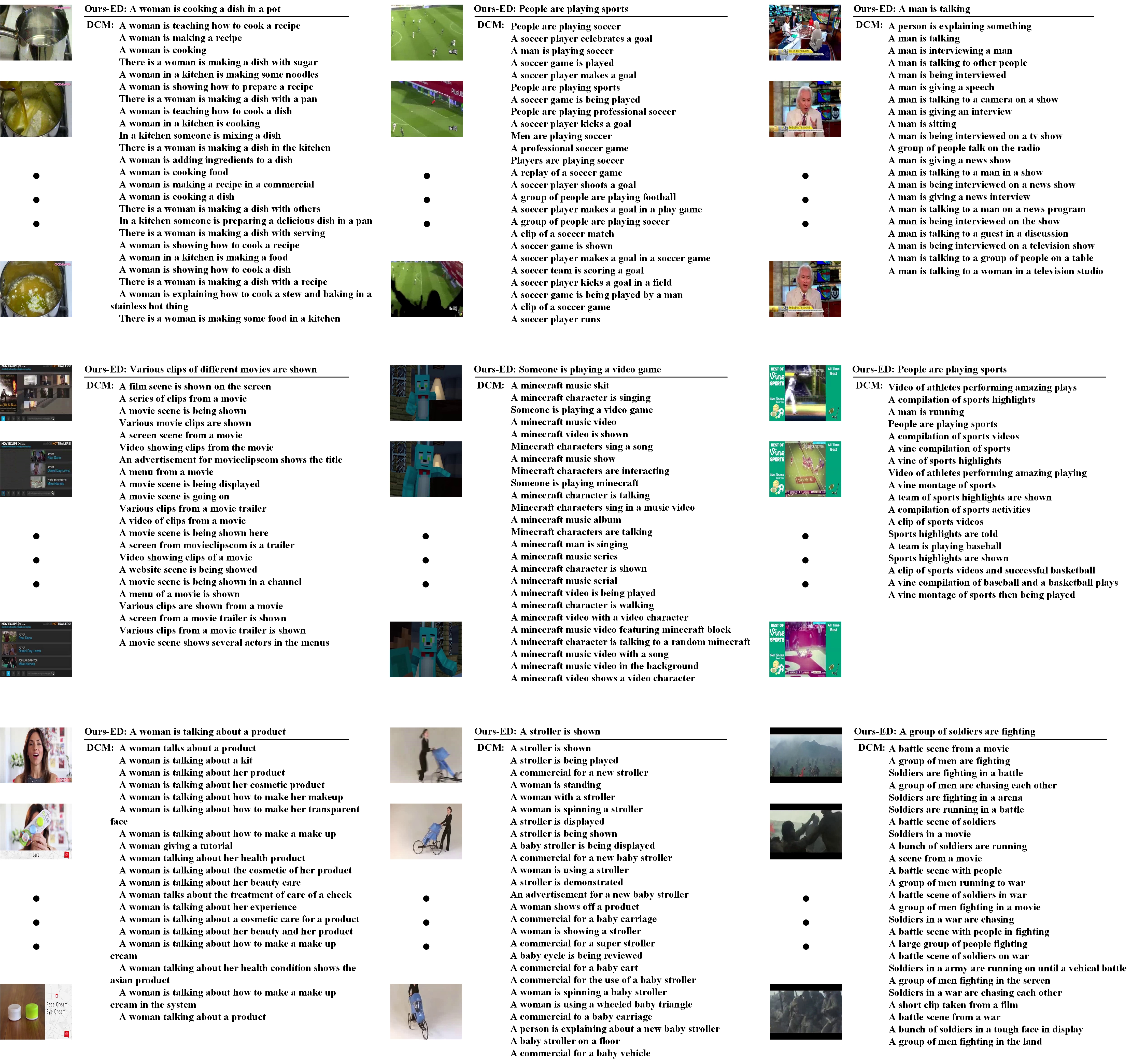}
	\caption{Examples of video captioning achieved by our model.}.
	\label{fig:picture_new_split1}
\end{figure*}

\subsubsection{Qualitative Analysis}
In our experiments, a beam search with size 5 is used to generate descriptions. Thus, we can obtain 5 sentences with each input \textbf{z}. We run the testing 10 times and remove duplicate sentences. To gain an intuition of the improvement on generated diverse descriptions of our DCM, we present some video examples with the video description from MS-RNN~\cite{song2018deterministic} and DVC~\cite{shen2017weakly} as comparison to our system in Fig. 3. We can see that our DCM performs better than MS-RNN and DVC in generating diverse sentences. Meanwhile, compared with these models, DCM generates more accurate and detailed descriptions for diverse video topics. For example, in the last video, DVC simply produces ``a man is hitting a ping pong ball in a stadium", while our model can generate sentence like ``two men compete in a game of table tennis by vigorously volleying the ping pong ball over the net", which shows the superiority of our method. At the same time, our model correctly describes ``two players" or ``two men". Conditioned on latent variables, our DCM can generate diverse descriptions while maintaining the semantic relevance. In general, compared with these two models, the sentences generated by DCM are superior in terms of video content description and sentence style generation.

To better understand our proposed method, in Fig. 4 we show some captioning examples generated by our model. From the generated results we can see that both Ours-ED and DCM are able to capture the core video subjects. Surprisingly, for a same model, in addition to describing video content from different aspects, DCM can generate sentences in different voices. For example, for the top video in the right column, DCM generates ``a man is being interviewed on a news show" and ``a man is giving a news interview". The former describes the video content in a passive voice, while the latter uses the active voice, which is a significant improvement over previous research works~\cite{yu2016video,song2018deterministic,shen2017weakly} that only produce a single voice. Combining the characteristic of CGAN and the design of fully convolutional network, our model achieves superior performance in generating diverse descriptions.
\begin{figure}[t]
	\centering
	\includegraphics[width=\columnwidth]{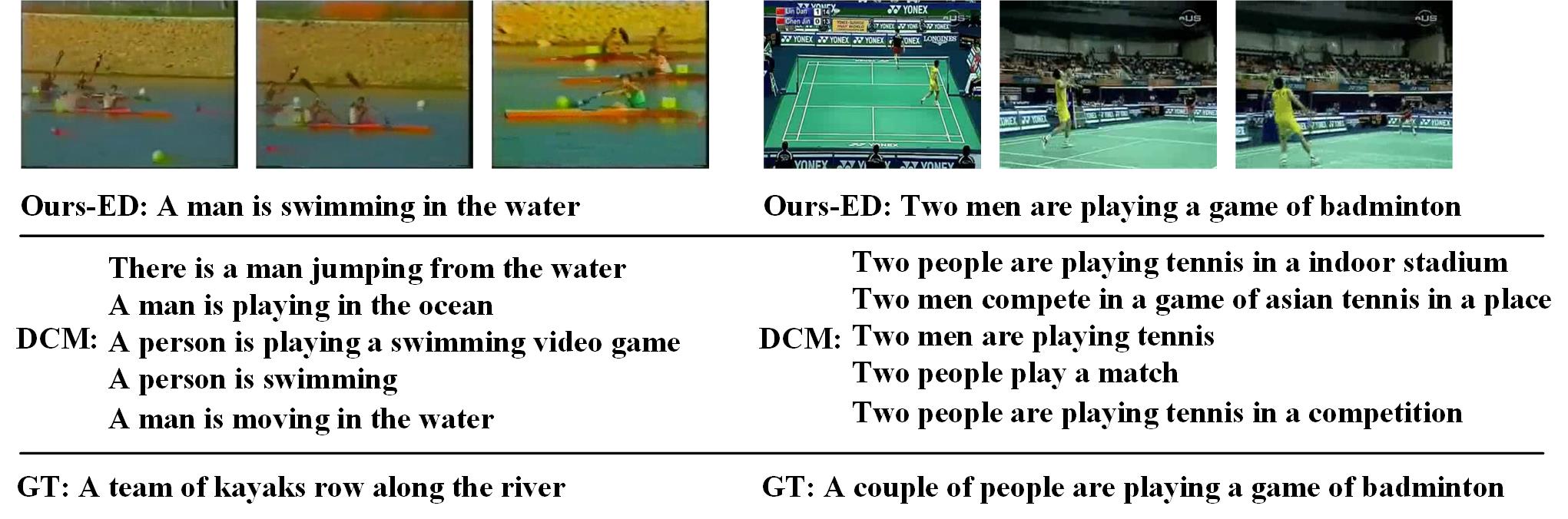}
	\caption{Failure cases of the description examples generated by our model.}.
	\label{fig:Failure cases of the description examples generated by our model}
\end{figure}

Although our model achieves satisfactory results, improvements can be made. Fig. 5 shows two failure cases, where our DCM approach fails to describe ``kayak" and ``badminton". For the first video, neither Ours-ED nor DCM can accurately describe ``kayak". This is due to the reason that our training dataset does not provide adequate training samples to distinguish this object. This indicates that to date, existing video captioning training datasets are still incomplete and require further refinement. For the second video, Ours-ED accurately describes ``badminton", while DCM fails. This is because when generating a description, in addition to relying on the latent variables extended from Ours-ED, DCM also relies on the generator of full convolution structure, which is less accurate than LSTM structure in generating natural language.
\begin{figure}[t]
	\centering
	\includegraphics[height=4cm,width=7cm]{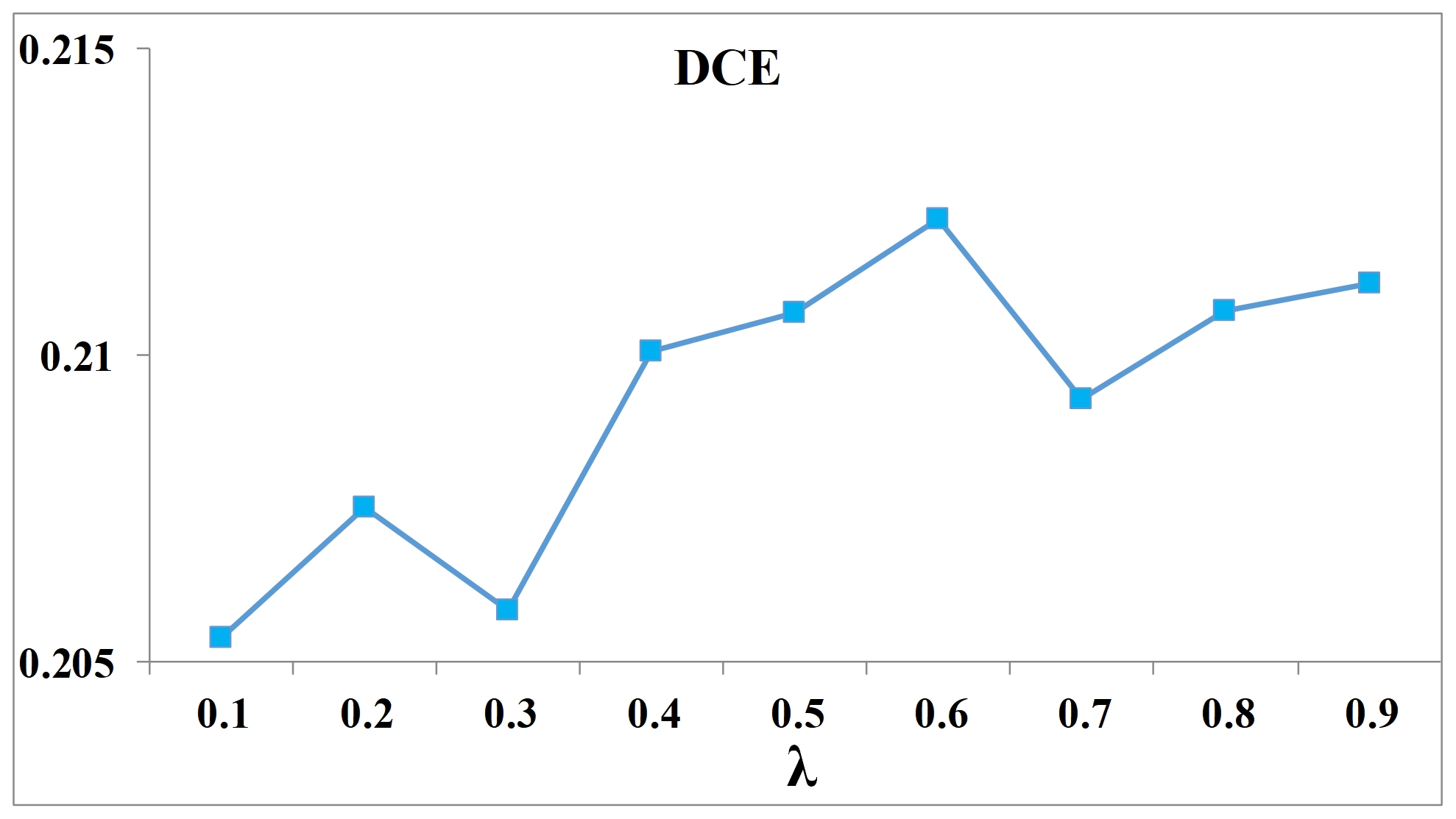}
	\caption{Effect of $\lambda$}
\end{figure}
\begin{figure}[t]
	\centering
	\includegraphics[height=4cm,width=7cm]{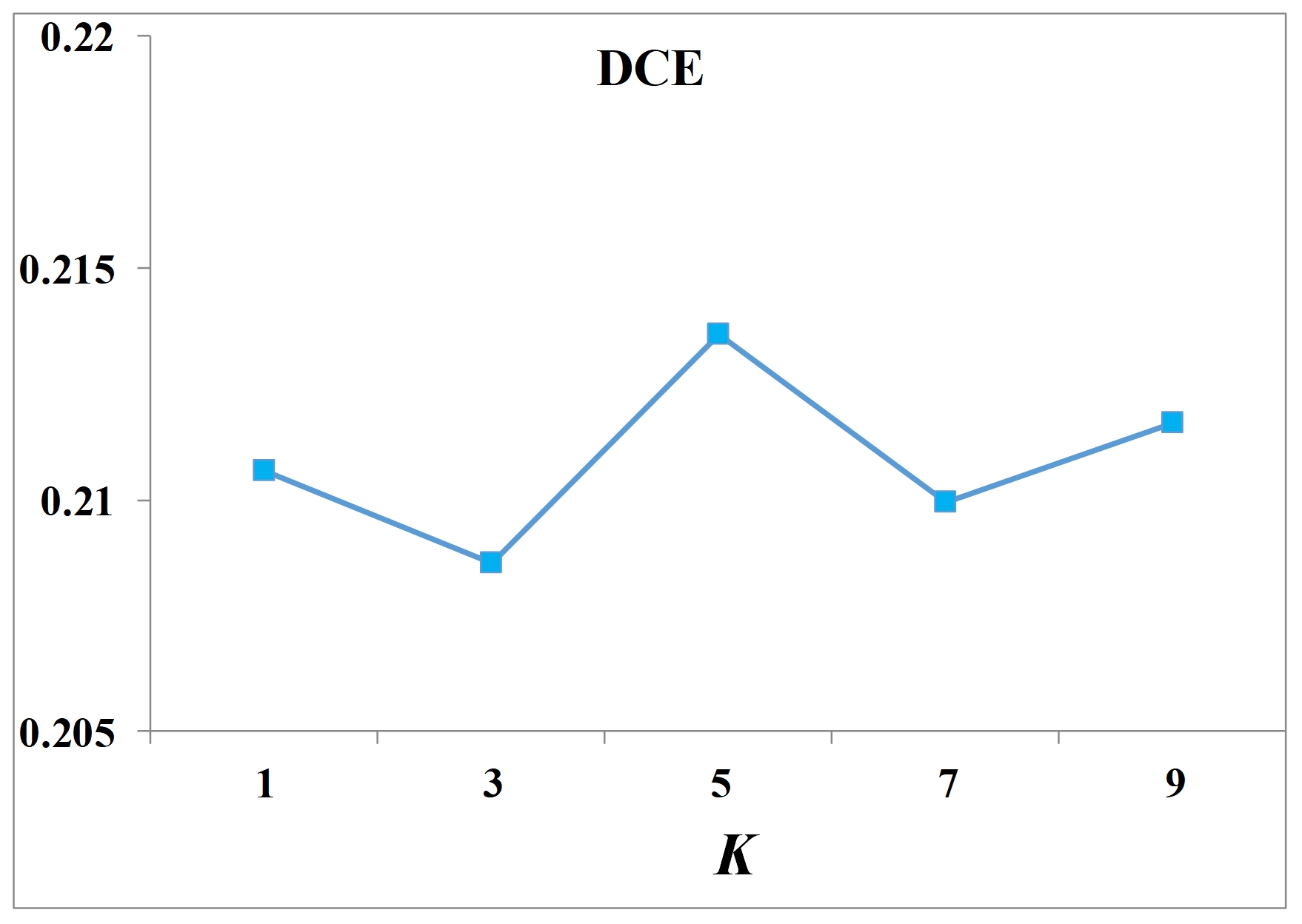}
	\caption{Effect of $K$}
\end{figure}
\begin{figure}[t]
	\centering
	\includegraphics[height=4cm,width=7cm]{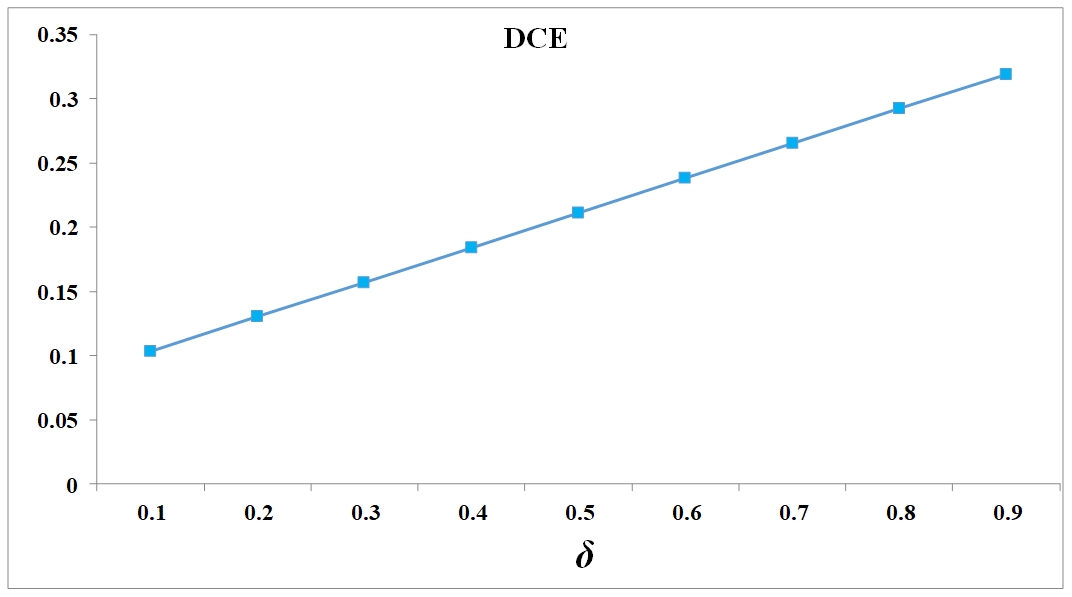}
	\caption{Effect of $\delta$}.
\end{figure}

\begin{table*}[t]
	\begin{center}
		\caption{Comparison of different architectures.} \label{tab:Comparison of different architecture}
		\begin{tabular}{c c c c c}
			\hline
			Model & Mixed mBLEU & Self-CIDEr & DCE-Variant & DCE\\
			\hline
			Single LSTM Layer& 0.831 & 0.665 & 0.297 & 0.188 \\
			Single CNN Layer& 0.772 & 0.744 & 0.343 & 0.192 \\
			Single LSTM Layer + Single CNN Layer& 0.830 & 0.673 & 0.304 & 0.194\\
			Single LSTM Layer + Multiple CNN Layers & 0.792 & 0.709 & 0.336 & 0.202\\
			Multiple CNN Layers & \textbf{0.687} & \textbf{0.803} & \textbf{0.387} & \textbf{0.210}\\
			\hline
		\end{tabular}
	\end{center}
\end{table*}

\subsection{Ablation Studies}
(1) \textbf{Effect of $\lambda$}. Considering both the diversity and fidelity of the generated sentences, and to speed up the convergence, we study the performance variance with different $\lambda$. We tune $\lambda$ from 0.1 to 0.9 on the MSR-VTT dataset. The results are reported in Fig. 6, from which we notice that when $\lambda=0.6$, our model achieves the highest DCE score. Thus we set $\lambda=0.6$ in the following experiments.

(2) \textbf{Effect of Rollout Times $K$}. Obviously, the larger $K$ is, the more time it costs. A suitable $K$ should improve the performance of the model without costing too much time. We increase the rollout times $K$ from 1 to 9 at intervals of 2 on MSR-VTT. Here we randomly select 10 sentences for each video and evaluate the diversity. As shown in Fig. 7, while we have $K=5$, our model achieves the best performance on DCE. Therefore, we adopt $K=5$.

(3) \textbf{Effect of $\delta$}. To investigate how $\delta$ affects DCE score, several ablation studies are conducted on MSR-VTT. Using Jaccard similarity coefficient for word-level diversity evaluation is a ``hard" method, while using BERT for sentence-level evaluation is relatively ``soft". For instance, for the sentences ``a woman is cooking" and ``a woman is showing how to cook a dish", their diversity (except accuracy) at the word level is 0.7, while the diversity at the sentence level is 0.12, which proves that the diversity evaluation at the sentence level is more stringent. As can be seen from Fig. 8, the larger the $\delta$, the greater the proportion of word-level evaluation, so the higher the DCE score.


(4) \textbf{Effect of Different Architectures of Generator}. To design a better diverse captioning model, we also conduct some ablation studies on different architectures of generator, as shown in Fig. 9. The details are as follow:

\begin{figure*}[t]
	\centering
	\includegraphics[width=\textwidth]{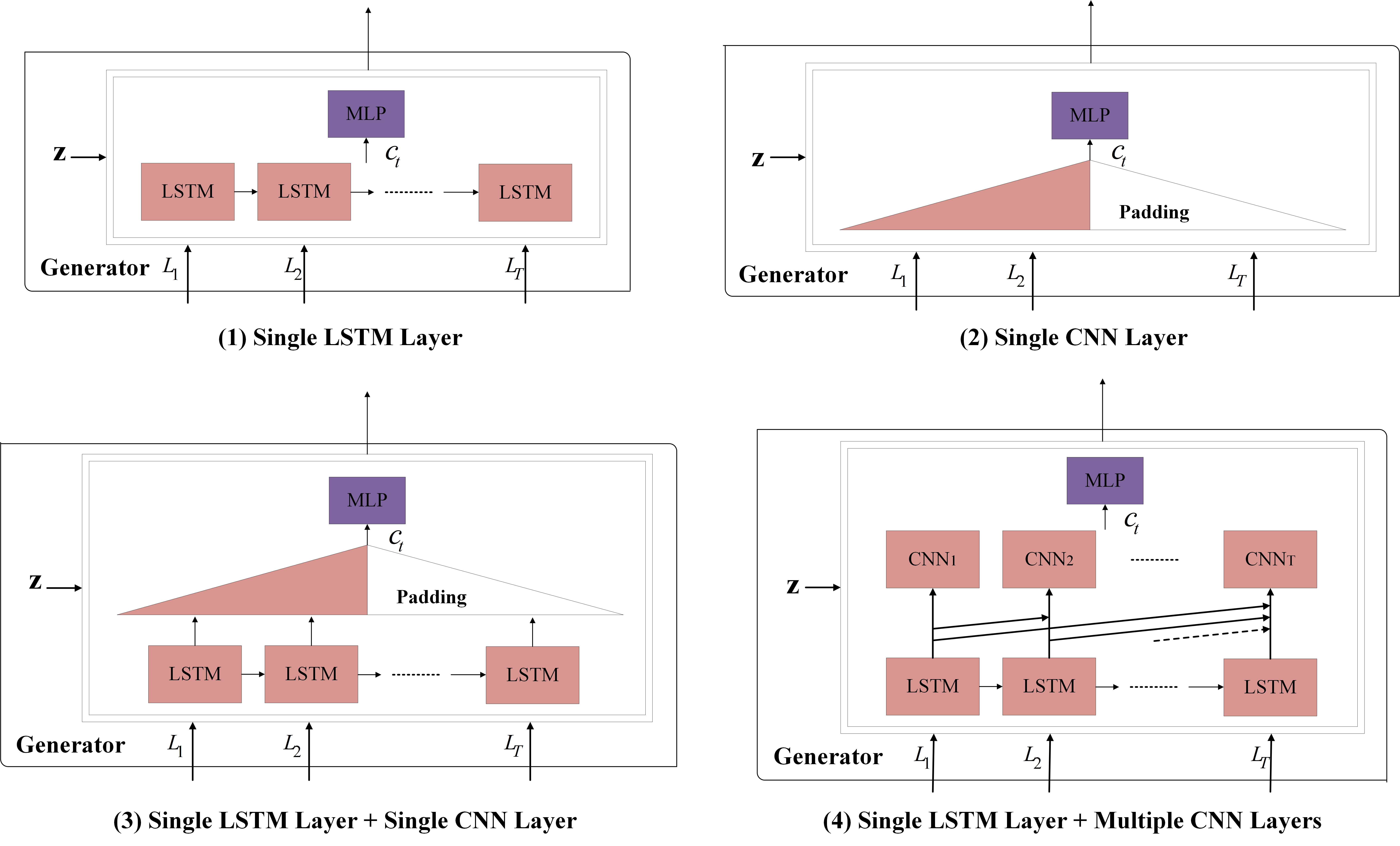}
	\caption{Different architectures of generator.}.
	\label{fig:different_architectures_of_generator}
\end{figure*}
\textbf{Single LSTM Layer}. Here a single LSTM is utilized as our generator to generate diverse captions. At each time step $t$, it directly takes $z_t$ and $L_t$ as input and produces the context vector $C_t$.

\textbf{Single CNN Layer}. Here a single CNN is utilized as our generator. At each time step $t$, suppose $s_t$ is the concatenation of $z_t$ and $L_t$, and $E_t\in \mathbb{R}^{T\times \bar{k}}=(s_1,...,s_t,pad,...,pad)$ is the concatenation of $s_{\leq t}$ and $(T-t)$ $pad$, where $pad$ is the zero vector with the same dimension as the $s_t$. Afterwards, the CNN with kernel $K_c^g$ $\in$ $\mathbb{R}^{T \times \bar{k}}$ takes input $E_t$ and produces the context vector $C_t$.

\textbf{Single LSTM layer + Single CNN layer}. Here we combine a LSTM and a CNN as our generator. At each time step $t$, the LSTM takes input $z_t$ and $L_t$, and outputs $q_t$ which is then combined with all the previous outputs $q_{<t}$ to produce a context vector $Q_t$, where $Q_t=\{q_1,...,q_t,pad,...,pad\}$ and $pad$ is the zero vector with the same dimension $k$ as the LSTM output. Afterwards, a CNN with kernel $K_c^g$ $\in$ $\mathbb{R}^{T \times k}$ is utilized to encode $Q_t$ and yields $C_t$.

\textbf{Single LSTM layer + Multiple CNN layers}. Here we combine a LSTM and multiple CNNs as our generator. At each time step $t$, the LSTM takes $z_t$ and $L_t$ as input and outputs $o_t$ which is then combined with all the previous outputs $o_{<t}$ to produce a context vector $O_t=(o_1,o_2,...,o_t)$. Afterwards, a ${\rm CNN_t}$ with kernel $K_c^g$ $\in$ $\mathbb{R}^{t \times k}$ is used to encode $O_t$ and produces $C_t$.

\textbf{Multiple CNN Layers}. This is our adopted method introduced above in the manuscript.

For each method we select the top 10 different sentences for each video and evaluate the diversity using Mixed mBLEU, Self-CIDEr, DCE-Variant and DCE. The results are presented in Table VI, showing that the architecture of multiple CNN layers performs best on all evaluation metrics. In particular, multiple CNN layers and single CNN layer achieve the top 2 performance in terms of Mixed mBLEU, Self-CIDEr and DCE-Variant, which demonstrates the potential of CNN architecture in generating diverse sentences. LSTM is good at producing peaky word probability distributions at the output~\cite{aneja2018convolutional}. This is conducive to generating high-precision descriptions, but limits the ability to generate diverse sentences. Compared to LSTM, CNN has the characteristic of generating lower peaky word probability distributions, giving it the potential to produce more word combinations. Benefit from the characteristics of CGAN, the diversity of single CNN architecture has been further improved, making it not only superior to the LSTM architecture, but also better than the combined architectures of LSTM and CNN on all metrics except DCE. Note that although the architectures of single LSTM layer + single CNN layer and single LSTM layer + multiple CNN layers are inferior to single CNN layer in terms of sentence diversity due to the LSTM limitations, they achieve better performance on DCE metric since LSTM makes the generated sentences more accurate, and DCE takes accuracy into account. For the last architecture, multiple CNNs are used for decoding. Each of them is responsible for predicting the next word based on a fixed length of input. Therefore, the division of labor of each CNN is very clear, and the information of all the previously generated words can be explicitly utilized when generating the next word, which is deficient in LSTM and single CNN. Compared to a single CNN, the architecture of multiple CNNs has advantages in generating high-precision sentences. In sum, it has a good performance in terms of both sentence diversity and accuracy. That is why it achieves the best results on all metrics. Thus, we adopt this multiple CNNs as our final architecture of generator.

\section{Conclusion}
In this work, a diverse captioning model DCM is proposed for video captioning, which simultaneously considers the fidelity and the diversity of descriptions. We obtain the latent variables that contain rich visual information and textual information from the well-designed encoder-decoder model, and then utilize them as input to a conditional GAN of full convolution design with the motivation to generate diverse sentences. Moreover, we develop a new evaluation metric named DCE to assess the diversity of a caption set. The potential of our method to generate diverse captions is demonstrated experimentally, through an elaborate experimental study involving two benchmark datasets MSVD and MSR-VTT.


%

\appendices




\ifCLASSOPTIONcaptionsoff
  \newpage
\fi

\bibliographystyle{IEEEtran}
\bibliography{IEEEexample}



\end{document}